%% file: iclr2027_conference.tex
\documentclass{article} % For LaTeX2e
\usepackage{preprint,times}

\usepackage{hyperref}
\usepackage{url}
\usepackage{cancel}
\usepackage{amssymb}
\usepackage{amsmath}
\usepackage{physics}
\usepackage{dsfont}
\usepackage{mathtools}
\usepackage{cleveref}
\usepackage{subcaption}
\usepackage{booktabs}
\usepackage{tabularx}
\usepackage{wrapfig}
\usepackage{algorithm}
\usepackage{algpseudocode}
\usepackage{tikz}
\usepackage{pgfplots}
\pgfplotsset{compat=1.18}
\usetikzlibrary{arrows.meta,backgrounds,external,fit,positioning,shapes.geometric}
\definecolor{linkblue}{HTML}{001473}
\input{notations.tex}

\hypersetup{
    colorlinks=true,
    linkcolor=linkblue,
    citecolor=linkblue,
    urlcolor=linkblue,
    pdfborder={0 0 0}
}
\usepackage{natbib}
\defcitealias{rozetScorebasedDataAssimilation2023}{SDA}
\defcitealias{shysheyaConditionalDiffusionModels2024}{Joint AR}
\defcitealias{savaryTrainingFreeBayesianFiltering}{Conditional AR}
\title{EnJoi: Ensemble Joint Score Filter\\ for Generative Data Assimilation}

\author{%
  Julien Moreau\\
  Inria Paris, DI ENS\\
  PSL Research University\\ Renault Group\\
  \texttt{julien.moreau@inria.fr} \\
  \And
  Marc Lelarge \\
  Inria Paris, DI ENS\\ PSL Research University\\
  \texttt{marc.lelarge@inria.fr} \\
}

\begin{document}

\maketitle

\begin{abstract}
Data Assimilation (DA) aims to recover the full state of a dynamical system that is only partially observed. A solution is to use Score-based models to generate physically consistent trajectories that agree with the observations. These Autoregressive Diffusion models are trained by conditioning on the previous state; however, they do not take into account the uncertainty of their past predictions. We propose a new diffusion-based assimilation algorithm that dynamically balances the confidence in the current state and the new observations. Crucially, we choose to learn the distribution of the \textit{joint} state containing both the past and future. This allows us to use a modified version of En4DVar, a classical DA algorithm that relies on the covariance of an ensemble of particles. Experiments on fluid and traffic flow simulations show improved reconstruction performance, especially in situations where observations are sparse and non-homogeneous. 
\end{abstract}

\section{Introduction}
Real world dynamical systems are often partially observed, making it challenging to estimate the full state at a given time. Scientists use numerical simulators to faithfully reproduce the dynamics, but they require an accurate starting point. Even with faithful initial conditions, the approximations in the physical model eventually lead to drift. A solution is to correct the simulation errors iteratively by ingesting sparse measurements; this practice is known as Data Assimilation (DA).
%  In this paper we only consider sequential filtering methods and simply refer to them as DA.
The framework has applications across diverse domains, from weather prediction \citep{carrassiDataAssimilationGeosciences2018} to traffic engineering \citep{workTrafficModelVelocity2010}.

DA problems are traditionally solved using Bayesian Filtering methods, applied to high-dimensional and non-linear systems. A classical algorithm is the Particle Filter (PF), which is asymptotically exact, but is notoriously subject to distribution collapse in large dimensions. Gaussian-linear methods (e.g.\ Kalman, 4DVar) are more robust but suffer from the linearization and mean-field assumption, resulting in overly smooth reconstructions. Modern systems often rely on a combination of the particle and Kalman formulation (i.e.\ LETKF \citep{huntEfficientDataAssimilation2007}, En4DVar \citep{claytonOperationalImplementationHybrid2013}). However, the accuracy of the filtering depends on computational bottlenecks in the number of particles and scale of the discretized simulation.

Fortunately, Deep-Learning surrogates offer a compute efficient alternative to traditional solvers. Learning to reconstruct the state given a history of measurements would require an unreasonable amount of data. Thus, neural simulators used for DA are trained on windows of few timesteps and learn a Markovian model of the dynamics. Because the state information is not sufficient to get a deterministic prediction of the future timesteps, this work will focus on generative emulators that learn a distribution of the possible states. 

Most generative DA methods train the surrogate like a standard \textit{conditional} forecast model, i.e.\ they expect the full past state as input \citep{huangDiffDADiffusionModel2024}. In this work, we argue that this conditioning discards the uncertainty of the past distribution (Section~\ref{section-limitations}). We therefore adopt a \textit{joint} model that learns to generate both the past and future states. Our main contribution is a new diffusion-based assimilation algorithm that combines the score of the joint state with the En4DVar framework (Section~\ref{section-algorithm}). The solution uses an ensemble of particles, which offers a more robust a priori of the past. The superiority of the method is demonstrated on three challenging assimilation experiments (Section~\ref{section-experiments}).

\paragraph{Notations} We study the state of a physical system $\x \in \mathbb{R}^d$ with observations $\y \in \mathbb{R}^{d_y}$. The superscript $k\in[0,T]$ denotes the time index, with $T$ the trajectory length. When considering an ensemble of size $N$, each particle is indexed with the $(i)$ superscript, where $i\in[\![1,N]\!]$. Given a symmetric positive definite matrix $\mathbf{M} \in \mathcal{S}^{d}_{++}$, we define the Mahalanobis norm as $\|\x\|^2_{\mathbf{M}} = \x^{\top} \mathbf{M}^{-1} \x$. Finally, to denote an intermediate state of the diffusion process, we use the subscript $\xt$, where $t$ is the noise level.

\section{Data Assimilation and filtering}

\paragraph{Dynamics and observation}
We consider a nonlinear state-space with forward model $\mathcal{A}:~\mathbb{R}^d~\to~\mathbb{R}^d$ and linear observation matrix $\mathbf{H} \in \mathbb{R}^{d_y\times d}$. The error terms $\boldsymbol{\epsilon}^k \in \mathbb{R}^d$, $\boldsymbol{\eta}^k \in \mathbb{R}^{d_y}$ are assumed to be independent, $\mathbf{Q}$ and $\mathbf{R}$ are the model and observation error covariances, respectively. 
\begin{align}
    \xkp &= \mathcal{A}(\xk) + \boldsymbol{\epsilon}^{k} 
    & \boldsymbol{\epsilon}^{k} &\sim \mathcal{N}(\mathbf{0},\mathbf{Q}) \label{eq:ssm-dynamics}\\
    \ykp &= \mathbf{H}\xkp + \boldsymbol{\eta}^{k+1},
    & \boldsymbol{\eta}^{k+1} &\sim \mathcal{N}(\mathbf{0},\mathbf{R}), \label{eq:ssm-observations}
\end{align}
The goal of online data assimilation or filtering is to recover the distribution of the true state from previous observations $p(\xkp|\yonekp)$.
% The \textit{background} distribution may consist in a single state estimate $\xk$, or an ensemble of $\xki$, $1\leq i \leq N$. Thus, we are often practically interested in how to propagate a single state while ingesting a single observation: $p(\xkp|\xk,\ykp)$.

\paragraph{Assimilation with 4DVar} Let us assume that we have access to the solution of the previous filtering step $p(\xk|\yonek)$, which we call the \textit{background} distribution. At time $k+1$, we look for the expected future state $\hatxkp:=\mathbb{E}[\xkp|\yonekp]$. When $\mathcal{A}$ is linear, estimating $\hatxkp$ is equivalent to finding the Maximum a posteriori (MAP) of the joint variable $(\xk,\xkp|\yonekp)$:
\begin{align}
    \checkxk, \hatxkp &= \underset{\xk,\xkp}{\arg \max } \quad p(\ykp|\xkp)p(\xkp|\xk)p(\xk|\y^{1:k}) \label{eq:4DVar}
\end{align}
where $\checkxk:=\mathbb{E}[\xk|\yonekp]$ denotes the smoothed past. Since we only need the most recent estimate $\hatxkp$, we can discard the smoothed variable $\checkxk$. For simplicity, we assume that the background is Gaussian, $p(\xk|\yonek)=\mathcal{N}(\xk|\hatxk,\mathbf{P})$, with constant covariance $\mathbf{P}$. In the general nonlinear case, an approximate MAP solution to \eqref{eq:4DVar} is obtained by differentiating the log-probability and applying iterative convex optimization \citep{Sanz-Alonso_Stuart_Taeb_2023}. The weak-constraint \textbf{4DVar} assimilation minimizes the following loss:
\begin{align}
    J_{\text{4DVar}}(\xk,\xkp) = \frac{1}{2}\|\xkp-\mathcal{A}(\xk)\|^2_{\mathbf{Q}} + \frac{1}{2}\|\xk - \hatxk\|^2_{\mathbf{P}} + \frac{1}{2}\|\ykp - \mathbf{H}\xkp\|^2_{\mathbf{R}} \label{eq:4DVar-loss}
\end{align}

\paragraph{Ensemble filtering}
In most real systems however, the background variance $\mathbf{P}$ is not stationary. An improvement is the \textbf{En4DVar} algorithm, which uses an ensemble of $N$ particles $\xki$ where $i \in [\![1,N]\!]$. For each particle, we solve a modified 4DVar objective where $\mathbf{P}$ is replaced by the empirical covariance $\mathbf{P}^k=\frac{1}{N}\sum_{i=1}^N (\xki - \barxk)(\xki - \barxk)^\top$, where $\barx^k$ is the empirical mean.
\begin{align}
    \checkxki, \xkpi &:= \underset{\xkkp}{\arg \max } \quad p(\ykp|\xkp)p(\xkp|\xk)\mathcal{N}(\xk|\xki,\Pk), \quad \forall i\in [\![1,N]\!] \label{eq:En4DVar}
\end{align}
Note that the background doesn't use the ensemble mean $\barxk$, but the particle $\xki$ to avoid ensemble collapse. The output of the ensemble filter is the empirical distribution $p(\xk|\yonek) \approx \frac{1}{N}\sum^N_{i=1} \delta(\xk-\xki)$.

\section{Score-based data assimilation}

\subsection{Posterior Sampling with Diffusion models}

\paragraph{Denoising Score Matching} Diffusion models are a generative method to sample from a target data distribution $\xo \sim p_{\text {data}}(\xo)$ starting from initial Gaussian noise $\x_1 \sim \mathcal{N}(\x_1;\mathbf{0},\mathbf{I})$, where $(\xo,\x_1) \in \mathbb{R}^{d} \times \mathbb{R}^d$. Let $t \in [0,1]$, the random variable $\x_t$ follows the Gaussian probability path $\x_t=a_t\x_0 + b_t\x_1$ where $a_t,b_t$ are the scale and noise coefficients in $\mathbb{R}$. At all times, we have the relation $\x_t|\x_0 \sim \mathcal{N}(\x_t\;|\;a_t \x_{0}, b_t^2 \mathbf{I})$. For conciseness, we omit the details of the training and generation process and refer the interested reader to Appendix~\ref{appendix-diffusion} or to the original papers for details \citep{hoDenoisingDiffusionProbabilistic2020,songScoreBasedGenerativeModeling2021}. We now assume that we have learned an approximation of the score function $\mathbf{s}_\theta(\xt,t) \approx \nabla_{\xt} \log p(\xt)$ from a dataset of physical states $\xo \sim p_{\text{data}}$.

\paragraph{Observation Guidance} Let us return to the observation equation \eqref{eq:ssm-observations}. The following method is valid for any timestep, thus we omit the $k$ in our notation. Given the likelihood $p(\y|\xo) = \mathcal N(\y|\mathbf{H}\xo,\mathbf{R})$, we are interested in generating \textit{posterior samples} $p(\xo|\y)$ that agree with the observations. This is an example of \textit{inverse problem} solving, which has been extensively studied in both diffusion and assimilation literature.
A popular solution is Diffusion Posterior Sampling (DPS) \citep{chungDiffusionPosteriorSampling2024}, a method to condition the generation of score-based models given a differentiable likelihood function $p(\y|\xo)$. Interestingly, the conditioning is performed at inference time, without any training on $\y$. DPS adds a \textit{guidance} term to the score function:
\begin{align}
    \nabla_{\xt} \log p(\xt|\y) &\approx s_\theta(\xt,t) - \frac{1}{2}\nabla_{\xt}\|\y-\mathbf{H}\tildexo(\xt)\|^2_{\mathbf{\tilde{R}}}
    \label{eq:dps}
\end{align}
Where we approximated $\xo$ with Tweedie's denoising expectation $\tildexo(\xt)= \mathbb{E}[\xo|\xt]\approx\frac{1}{a_t}(\xt + b_t^2 s_\theta(\xt,t))$ and the gradient is computed with auto-differentiation. The variance $\mathbf{\tilde R}$ is chosen as a function of the observation error $\mathbf{R}$ (see Appendix~\ref{appendix-dps} for details). It is common practice to treat $\mathbf{R}$ as a hyperparameter and to choose a different value from the real observation noise $\mathbf{R}_{\text{true}}$.

\subsection{Trajectory sampling}

Until now, we have reviewed generative models that ignore the temporal dynamics from the state space model. For trajectory generation, we can train our score-based model on a dataset of state pairs $(\xk,\xkp)$ \footnote{Note that for simplification we focus on 2-step windows but the state $\x$ can be chosen as a concatenation of several time-lags for additional context.}. At inference time, by letting the model access the previously generated state, we can iteratively generate a rollout of any length. In the literature, two architectures have been proposed to obtain an autoregressive (AR) diffusion. The model can either be trained to generate only the future state $\xkp$ \textit{conditioned} on the past $\xk$, or to generate a pair of \textit{joint} states $\xkkp$. The two modalities are usually regarded as equivalent. 
% At each k, we can apply diffusion posterior sampling for reconstruction
% Conditional is used in this setting
% Joint is used in this setting
\paragraph{Conditional AR} The most common practice is to learn the \textit{conditional} score $s_\theta(\xkpt|\xk)$, where a clean background state $\xk$ is fed as additional input to the score network \citep{savaryTrainingFreeBayesianFiltering}. The observations can be ingested with DPS guidance \eqref{eq:dps}:
\begin{equation}
    \nabla_{\xkpt} \log p(\xkpt|\xk,\ykp) \approx \underbrace{s_\theta(\xkpt|\xk,t)}_{\text{conditional score}} - \underbrace{\frac{1}{2}\nabla_{\xkpt} \|\ykp-\mathbf H\tildexkpo(\xkpt|\xk)\|^2_{\mathbf{\tilde R}}}_{\text{guidance on observation }k+1}  \label{eq:cond-sda}
\end{equation}

\paragraph{Joint AR} The other possibility is to denoise both timesteps jointly. An early version of this joint score was proposed in SDA \citep{rozetScorebasedDataAssimilation2023} for smoothing, and then adapted to filtering by \citet{shysheyaConditionalDiffusionModels2024} using reconstruction guidance \citep{hoVideoDiffusionModels2022} to obtain a joint autoregressive model:
\begin{equation}
    \nabla \log p(\xkkpt|\xk,\ykp) \approx \underbrace{s_\theta(\xkkpt,t)}_{\text{joint score}} - \underbrace{\frac{1}{2}\nabla \|\xk - \tildexko\|^2_{\mathbf{\tilde P}}}_{\text{reconstruction guidance}} - \underbrace{\frac{1}{2}\nabla \|\ykp-\mathbf H\tildexkpo\|^2_{\mathbf{\tilde R}}}_{\text{guidance on observation }k+1}  \label{eq:joint-sda}
\end{equation}
For readability, we have omitted that the gradient is taken with respect to $\xkkpt$ and that the Tweedie expectations $\tildexko(\xkkpt)$ and $\tildexkpo(\xkkpt)$ depend on the state.
The reconstruction guidance ensures that at the end of denoising, $\xko \approx \xk$ with  variance $\mathbf{P}=\sigma_{\min}^2\mathbf{I}$ as penalty (the deviation $\sigma_{\min}$ is chosen as a small value). Additional motivation and sampling variants are available in Appendix~\ref{appendix-scores}.
% Differences between this formulation and approaches from the litterature are highlighted in Table~\ref{tab:trajectory-sampling},

\paragraph{Prediction and correction}
Notice that both approaches view the previous state $\xk$ as exact conditioning, not as a background distribution. \citet{shysheyaConditionalDiffusionModels2024} show that the joint formulation is more flexible but that a conditional score is generally more efficient for long rollouts (without observations). However, if the state estimation $\xk$ is wrong and rich observations $\ykp$ are available, the joint model will be better at \textit{correcting} itself, because it is only weakly constrained by the reconstruction guidance. For conditional models, the correction mechanism has to be emulated by using an ensemble of particles and eliminating less probable states, as illustrated in recent studies \citep{savaryTrainingFreeBayesianFiltering,weiSURGEApproximationfreeTraining2026}. Note that another research direction is to learn a \textit{static} model of a single timestep, and to rely on the numerical simulator for the dynamics \citep{andraeDAISIDataAssimilation2026}. More details on related works are in Appendix~\ref{appendix-related-works}.

\section{Current Models Failure modes}
\label{section-limitations}
% TODO rewrite in this order

% Main claims:
% 0 - Conditional models are slow to correct themselves when starting from a wrong point
% Show it with a warmup experiment

% 1 - Even the joint "reconstruction" models are too confident on the past, thus making them more "volatile" by reducing the guidance $R$ can be beneficial.

% Prove it by showing that volatile models are better in experiments (even with large observation noise). However, in scenarios where observations are numerous and low quality, they shouldn't be very good. I could increase the Bernoulli ratio in the sparse experiments to show it

% 2 - Explain that chosing a constant P,R is not the most practical. Introduce the scaled guidance ? 

In this section, we identify the limitations of current filtering algorithms. We argue that the existing methods were designed for scenarios with perfect initial condition, low noise and homogeneous observations. Thus, they are not robust to more difficult settings encountered in real systems. We propose to study three toy scenarios on 1D Kuramoto-Sivashinsky fluid data to highlight the challenges of generative data assimilation. The algorithms are run with 10 particles and the performance is evaluated with CRPS, which measures the error of the ensemble empirical distribution (see Appendix~\ref{appendix-metrics} for the metrics definition).

\begin{figure}[!htb]
    \centering
    \includegraphics{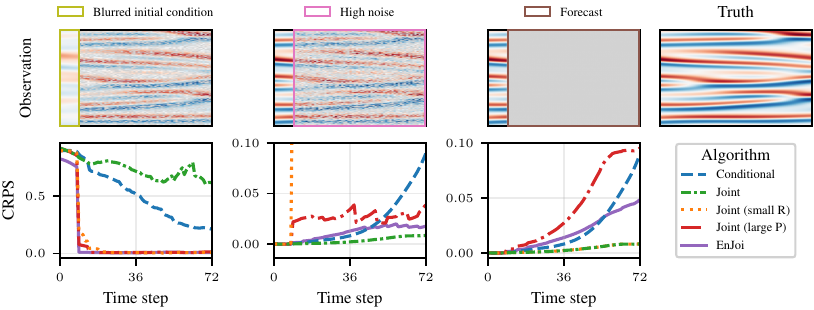}
    \caption{Three assimilation scenarios on 1D Kuramoto Sivashinsky data. \textbf{Left: Scenario 1 (Warmup)}, the initial condition is perturbed and the filter must correct itself with the observations. \textbf{Center: Scenario 2 (Noise)}, from perfect initial state, the filter should assimilate noisy measurements. \textbf{Right: Scenario 3 (Forecast)}, the model must predict the state without observations. Our proposed EnJoi Score Filter offers the best compromise. We note that our method isn't the most efficient to forecast from perfect initial conditions, but we argue that this is not an issue in the assimilation setting.}
    \label{fig:ks-combined}
\end{figure}

\paragraph{Correcting an imperfect background}
At the beginning of the trajectory, existing data assimilation benchmarks assume that the filters are initialized with a good first estimate of the state. However, it is common for Kalman Filters to start with a rough estimate and slowly increase the fidelity of the reconstruction by ingesting observations. We study this ability to warm-up and stabilize in \textbf{Scenario 1 (Warmup)}, the initial condition is blurred and the model must correct the state with sparse measurements. Each cell is observed with Bernoulli probability $p=0.5$ and the observation noise is set to $\sigma_{\text{obs}}=0.1$. In Figure~\ref{fig:ks-combined} (left), we observe that the methods relying on exact conditioning (Joint and Conditional) have a large inertia and fail to correct their trajectory. A solution is to modify the guidance hyperparameters: either increase P or decrease R. We obtain a \textit{volatile} model that quickly adapts to incoming observations. This correction mechanism is crucial when observations are more informative than the background, notably when the estimated state has drifted from the true trajectory.

\paragraph{Balancing observation and background}
We have seen that current models are too stiff to rapidly correct their trajectory. However, this inertia can be beneficial when confronted with noisy observations. Modifying the guidance to obtain a \textit{volatile} model (e.g. by choosing a small R) will cause the filter to be overly sensitive to measurements. In \textbf{Scenario 2 (Noise)}, we start from perfect initial conditions, use the same Bernoulli observations ($p=0.5$) but increase the observation noise to $\sigma_{\text{obs}}=0.5$. As seen in Figure~\ref{fig:ks-combined} (center), the \textit{volatile} models are not robust to the increased noise. The model with small $R$ inherits the errors of the measurements and the score network is unable to produce a smooth physical state. In \textbf{Scenario 3 (Forecast)}, we let the algorithms roll out their prediction from perfect initial conditions without any observation. In Figure~\ref{fig:ks-combined} (right), we observe that choosing a larger P causes the Joint model to quickly forget the background distribution and to drift from the true trajectory. These experiments show that a correct equilibrium must be found between the confidence on the past, trust in the physical model and sensitivity to observations. When the uncertainty is stable in time, it is common to use $\mathbf{P}$ and $\mathbf{R}$ that are constant across the trajectory length and the state dimension, but this is not always a good approximation in real settings.

\paragraph{Spatial uncertainty}
Once the correct tradeoff between observation and past has been found, there is still the issue of spatially-localized uncertainty. In real systems, it is common to have good coverage of some regions and poor information on others. In the case of moving sensors, data loss or missing values, the uncertainty pattern can evolve in space-time. It is also common to have highly chaotic regions with higher uncertainty due to the dynamics of the system (e.g. wave fronts, attractors). This spatial uncertainty is not incorporated into current diffusion algorithms: the joint model from \citet{shysheyaConditionalDiffusionModels2024} uses a constant diagonal guidance, while particle-filter methods compute likelihood weights for the entire state-space at once.

These experiments motivate that our current approximation of the background distribution is not flexible enough for the assimilation setting. The exact conditioning hypothesis will not hold for long rollouts, the constant confidence in the past state is a simplification and the spatial dimension is not taken into account.

\section{Proposed Method}
\label{section-algorithm}
% \subsection{Including a background probability}
Modern data assimilation algorithms use an ensemble of particles to account for the diversity of the filtering distribution. We propose to adapt the En4DVar \eqref{eq:En4DVar} framework, which maintains an expressive background prior by computing the covariance of previous predictions. Instead of optimizing the state by differentiating the numerical model \eqref{eq:4DVar-loss}, we will use a joint generative model.

\subsection{Guidance with ensemble covariance}

We first need to adapt 4DVar \eqref{eq:4DVar} to a setting with diffusion noise. This can be achieved with a series of approximations to extract one guidance for the past and one for the future.
\begin{align}
    p(\xkkpt|\yonekp)&\propto p(\xkkpt)p(\yonekp|\xkkpt)\notag \\
        & \approx p(\xkkpt)\overbrace{p(\yonek|\xkkpt)p(\ykp|\xkkpt)}^{\approx \text{independent}}\notag \\
        & \propto p(\xkkpt)\overbrace{p(\xkkpt|\yonek)}^{\approx \text{use prior as likelihood}}p(\ykp|\xkkpt) \notag \\
        & \underset{\text{DPS}}{\approx} p(\xkkpt)\overbrace{\mathcal{N}(\tildexko|\hatxk,\Pk)}^{\text{Kalman recurrence}}\mathcal{N}(\ykp|\mathbf{H}\tildexkpo,\mathbf{R}) \label{eq:two-guidances}
   %//  \nabla_{\xkkpt} \log p(\xkkpt|\yonekp) &\approx s_\theta(\xkkpt) - \underbrace{\frac{1}{2}\nabla_{\xkkpt} \|\hatxk - \tildexko\|^2_{\Pk}}_{\text{background guidance}} \notag \\ &\quad - \underbrace{\frac{1}{2}\nabla_{\xkkpt} \|\ykp-H\tildexkpo\|^2_{R}}_{\text{observation guidance}} \label{eq:naive-guidance}
\end{align}
Where we have used DPS to replace the state $\xkt$ with $\tildexko$. Note that the Gaussian background prior acts as a likelihood. This is for practical considerations and is further motivated in Appendix~\ref{appendix-prior-swap}.
We can then derive a score that is similar to Joint AR \eqref{eq:joint-sda}, but the guidance strength is now dependent on the ensemble covariance $\Pk$.

\paragraph{An issue when guidance is small}
Equation \eqref{eq:two-guidances} allows us to approximately sample from the filtering distribution. However this is not enough to fulfill our objective, since the En4DVar algorithm \eqref{eq:En4DVar} requires the MAP. With posterior sampling, there is no guarantee that a sample will be from the most probable mode. If we are in a long term rollout, $\Pk$ will grow, and if no observation is available, the observation guidance will vanish (infinite $\mathbf{R}$). In this situation with low guidance signal, the diffusion model will generate unconditional samples and the filter will tend to forget its past.

\subsection{Scaling the Guidance}

% \paragraph{Inflate prior guidance ?}
% Now if the problem is only one of vanishing guidance, there is a straight-forward solution which is to maintain strong guidance at all times. The easy option is to rescale the covariance $\Pk$ or to replace it with a constant value. Thus we get back to the reconstruction framework \ref{eq:rec-sda}.

% \paragraph{Inflate both guidances ?}
% One issue of inflating the auto-regressive term is that it will hinder the observation guidance. One could alternatively inflate both terms, which will produce unphysical solutions when the background and observations disagree. We will see that this is not a problem in practice, however the magnitude of observation error is now hard to inject into the model.

\paragraph{Balancing guidance terms}
To maintain strong guidance at all times, we propose to scale both guidance terms together. This is a way to preserve the balance between observation noise and past uncertainty. We rewrite the noisy MAP problem as the sum of two conflicting objectives:
\begin{align*}
    \underset{\xkkpt}{\arg \max} \quad \underbrace{\log p(\xkkpt)}_{\text{consistency}} + \underbrace{\log \mathcal{N}(\tildexko|\hatxk,\mathbf{P}^k)\mathcal{N}(\ykp|\mathbf{H}\tildexkpo,\mathbf{R})}_{\approx -  J^\star_{\text {4DVar}}(\tilde{\x}^{k:k+1}_0)}
\end{align*}
Where the diffusion score is acting as a regularizer, pushing the noisy estimates towards solutions that are consistent with both the physics and the noise level. The second term can be interpreted as the 4DVar objective \eqref{eq:4DVar-loss} without a physical model $(\mathbf{Q}=+\boldsymbol{\infty})$.
We normalize the guidance term with the root mean square to ensure that the two objectives are balanced.
\begin{align}
    \mathbf{s} &= \nabla_{\xkkpt} \log p(\xkkpt) - \gamma \sqrt{2d} \frac{\nabla_{\xkkpt} J^\star_{\text{4DVar}}(\tilde{\x}^{k:k+1}_0)}{\left\|\nabla_{\xkkpt} J^\star_{\text{4DVar}}(\tilde{\x}^{k:k+1}_0)\right\|_2} \label{eq:scaled-guidance}
\end{align}
Where $2d$ is the dimension of the joint state $\xkkpt$ and we introduce the guidance strength hyperparameter $\gamma$ which we set to 1. Similar to \citet{rozetScorebasedDataAssimilation2023}, we apply an additional step of Langevin sampling after the SDE integration step, to correct the approximation of the guidance signal.
\begin{align}
    \xkkpt &\gets \xkkpt + \tau \cdot \frac{\mathbf{s}}{\|\mathbf{s}\|_2} + \sqrt{\tau} \mathbf{\epsilon_c}, \quad \epsilon_c \sim \mathcal{N}(\mathbf{0},\mathbf{I}), \quad \tau \in \mathbb{R}
    \label{eq:langevin}
\end{align}
% A potential drawback is that the scaled gradient could make the descent oscillate indefinitely. However we are using diffusion models that denoises the target with a step proportional to $b_t \nabla \log p$. Thus the stepsize naturally vanishes when $t \to 0$. 

\paragraph{Related Guidance methods} Rescaling the guidance and keeping only directional information has been proposed in other works. In particular, Spherical Guidance \citep{yangGuidanceSphericalGaussian2024} proposes to project the state to the sphere of center $\mathbb{E}[\xtm|\xt]$ with a radius proportional to $\sqrt{d}$ to stay in the denoising manifold. Later work DMAP \citep{xuRethinkingDiffusionPosterior2025} adopts an optimization perspective and proposes to alternate spherical projection with DPS steps. Their algorithms are very close to our scaled guidance with Langevin corrections.

\paragraph{Maintaining ensemble spread} Because of the random noise initialization, the model will generate various scenarios in the areas where the guidance is not constraining the generation. If there are multiple local maxima in the posterior distribution, using the MAP formulation will not make the filter collapse to a single particle. This bears resemblance to a stochastic En4DVar framework where model noise $\boldsymbol{\epsilon} \sim \mathcal{N}(\mathbf{0},\mathbf{Q})$ is injected into the simulation to increase ensemble spread.

\subsection{The Ensemble Joint Score Filter algorithm}

We refer to our method as Ensemble Joint Score Filter (\textit{Enjoi}), since it is the combination of En4DVar with a Joint score function. The iterative filtering process is described in Algorithm~\ref{alg:ensemble-diffusion-filter} which closely resembles the classical En4DVar algorithm. The key difference is that the approximate MAP is generated using a joint score neural network with our scaled guided sampling. We detail this modified generation process in Algorithm~\ref{alg:scaled-diffusion-sampling}.

Compared to other diffusion filters, the novelty of our approach is to incorporate the ensemble covariance into the sampling. This has several benefits, which stem from the En4DVar framework. Our algorithm takes full advantage of the \textbf{empirical past uncertainty}: the reconstruction chooses the right balance between the uncertain past and the noisy observation. This is an improvement over auto-regressive methods that set a constant confidence to the past state (e.g. using a specific noise level \citep{andraeDAISIDataAssimilation2026,jiaForcingDASUnifiedRobust2026} or guidance strength \citep{shysheyaConditionalDiffusionModels2024}). The other strength of our method is its ability to perform \textbf{local updates}, which alleviates the curse of dimensionality of Particle Filters \citep{savaryTrainingFreeBayesianFiltering,weiSURGEApproximationfreeTraining2026,baoScorebasedNonlinearFilter2023}. Particles that diverge from the MAP are locally course-corrected instead of being eliminated. Thus, we do not lose the entire state information. This allows the filter to track multiple sources of uncertainty with a reduced ensemble size.
    % \item \textbf{Sharp Generative prior} Similarly to En4DVar, we make the simplifying assumption that the background is a Multivariate Gaussian, thus we are not able to handle non-linear or multimodal past priors. However, the neural score function has learned the joint data distribution and generates sharp, physically sound samples, not overly smooth averages. % We avoid a common tendency of tangent-linear interpolation methods (e.g. Kalman, 4DVar) to generate overly smooth trajectories.

\begin{algorithm}[!htb]
\caption{EnJoi: Ensemble Joint Score Filter}
\label{alg:ensemble-diffusion-filter}
\begin{algorithmic}[1]
\Require Initial ensemble $\{\mathbf{x}^{k=0, (i)}\}_{i=1}^N$, observations $\{\ykp\}_{k=0}^{T-1}$, $\mathbf{P}^0$
\For{$k = 0, \ldots, T-1$} 
    \State $\xkpi \gets \operatorname{JointDiffusion}(\xki,\Pk,\ykp), \quad \forall i \in [\![1,N]\!]$ \Comment{Sample approximate MAP}
    \State $\barxkp \gets \frac{1}{N} \sum_{i=1}^N \xkpi$
    \State $\Pkp \gets \frac{1}{N} \sum_{i=1}^N (\xkpi - \barxkp)(\xkpi - \barxkp)^\top$ \Comment{Estimate background covariance}
    \State $\Pkp \gets \operatorname{GaspariCohn}(\Pkp) + \sigma_{\min}$ \Comment{Apply localization \citep{gaspari1999construction}}
\EndFor \\
\Return Empirical distribution $\{\mathbf{x}^{1:T,\,(i)}\}^{N}_{i=1}$ and covariance $\{\Pk\}^{T}_{k=1}$ 
\end{algorithmic}
\end{algorithm}

\begin{algorithm}[!htb]
    \caption{Guided Joint Diffusion Sampling}
    \label{alg:scaled-diffusion-sampling}
    \begin{algorithmic}[1]
        \Require Background state $\xki$, state covariance $\Pk$, observation $\ykp$, Observation covariance $R$, score $s_\theta$, correction stepsize $\tau$, guidance strength $\gamma$.
        \State $\xkkpt \sim \mathcal{N}(\mathbf{0},\mathbf{I})$ \Comment{Sample random noise}
        \For{$t = 1, 1-\Delta t, \ldots, \Delta t$}
            \State \label{alg:line:tweedie} $\tildexkkpo \gets \frac{1}{a_t}(\xkkpt + b_t^2 s_\theta(\xkkpt))$ \Comment{Approximate reconstruction with Tweedie}
            % \State $J_{\text {4DVar}}(\tildexkkpo) = \|\xki - \tildexko\|_{\Pk}+ \|\ykp - H\tildexkpo\|_{R}$
            \State \label{alg:line:guided-score} $\mathbf{s} \gets s_{\theta}(\xkkpt,t) - \gamma \sqrt{2d} \frac{\nabla_{\xkkpt} J^\star_{\text{4DVar}}(\tildexkkpo)}{\|\nabla_{\xkkpt} J^\star_{\text{4DVar}}(\tildexkkpo)\|}$ \Comment{Add guidance \eqref{eq:scaled-guidance} to the score}
            % \State $\xkkptm \gets \sqrt{ \alpha_{t-1} } \cdot \tildexkkpo - \sqrt{1 - \alpha_{t-1} - \sigma_{t}^2}  \sqrt{1-\alpha_t} \cdot \mathbf{s} + \sigma_{t} \cdot \boldsymbol{\epsilon}(t)$ % Stochastic
            % \State $\xkkptm \gets \sqrt{\frac{\alpha_{t-\Delta t}}{\alpha_{t}}} \cdot \xkkpt - (\sqrt{1-\alpha_{t-\Delta t}} - \sqrt{\frac{\alpha_{t-\Delta t}}{\alpha_{t}}}\sqrt{1 - \alpha_{t}}) \sqrt{ 1 -\alpha_t} \cdot \mathbf{s}$ 
            \State $\xkkptm \gets \operatorname{SDESolverStep}(\xkkpt,\mathbf{s},t)$ \Comment{Integrate backward SDE \citep{luDPMSolverFastODE2022}}
            % \For{$c = 1, \ldots, C$}
            % \State $\mathbf{s}_c \gets s_{\theta}(\xkkptm,t-\Delta t) - \gamma \sqrt{2d} \frac{\nabla_{\xkkptm} J_{\text{4DVar}}(\tildexkkpo)}{\|\nabla_{\xkkptm} J_{\text{4DVar}}(\tildexkkpo)\|}$ \Comment{Recompute guided score}
            % \State $\xkkptm \gets \xkkptm + \tau \cdot \mathbf{s}_c/\|\mathbf{s}_c\| + \sqrt{\tau} \mathbf{\epsilon_c}, \quad \epsilon_c \sim \mathcal{N}(\mathbf{0},\mathbf{I})$
            \State $\mathbf{s}' \gets$ recompute lines \ref{alg:line:tweedie}--\ref{alg:line:guided-score} at $t-\Delta t$ \Comment{Recompute the guided score}
            \State $\xkkptm \gets \operatorname{LangevinStep}(\xkkptm,\mathbf{s}',\tau)$ \Comment{Apply one Langevin correction \eqref{eq:langevin}}
            % \EndFor
        \EndFor \\
        \Return Denoised state $\xkpo$
    \end{algorithmic}
\end{algorithm}

\paragraph{Background Covariance} To avoid spurious correlations in the empirical covariance, we apply Gaspari-Cohn localization \citep{gaspari1999construction}. This classical DA method applies a spatio-temporal kernel function with bounded support to the covariance matrix. The localization radius must be chosen empirically considering the temporal and spatial scale of the dynamical process. A preconditioning diagonal term $\sigma_{\min}$ is also added for stability, we use the same value as the reconstruction guidance \eqref{eq:joint-sda}. The constant covariance term is meant to represent stationary or climatic variance. Operational DA systems often rely on \textit{hybrid} covariance matrix for robustness.

\section{Experiments}
\label{section-experiments}

We compare our \textbf{EnJoi} algorithm to two approaches from the recent literature. The first baseline consists of i.i.d.\ runs of Conditional AR Diffusion from \citet{shysheyaConditionalDiffusionModels2024}. We use their implementation of \textbf{Joint AR} \eqref{eq:joint-sda} and \textbf{Cond. AR} \eqref{eq:cond-sda}. For comparison with a recent ensemble method, we re-implement the particle filter from \citet{savaryTrainingFreeBayesianFiltering} which we refer to as \textbf{Cond. P.F.}. We also generalize the method to the joint score, resulting in a \textbf{Joint P.F.}.

Motivated by our experiments in Section~\ref{section-limitations}, we explore the possibility of selecting a smaller variance $R$ than the real measurement noise $\sigma_{\text{obs}}^2$. We denote this modification as \textit{volatile} (v) models. For clarity, the algorithms using the true variance $\sigma_{\text{obs}}^2$ are tagged with an (s) for \textit{steady}. We use \textit{(g=SDA)} to denote the guidance from \cite{rozetScorebasedDataAssimilation2023}, \textit{(g=scaled)} is our scaled guidance \eqref{eq:scaled-guidance} with constant $\mathbf{P}$, and \textit{(g=MMPS)} is the Moment-Matching Posterior Sampling from \cite{rozetLearningDiffusionPriors2024}. Unless specified otherwise, the number of particles is set to 10 for all algorithms.

\subsection{1D Kuramoto Sivashinsky}
% TODO: put this in appendix as well
\paragraph{Data} We use the KS dataset from \cite{shysheyaConditionalDiffusionModels2024}. This synthetic experiment considers a 1D fluid simulation with a chaotic behaviour, which makes the assimilation challenging. In the \textit{Bernoulli} setting, each cell of the space-time grid has probability $p=0.1$ to be observed. We experiment with different observation noise levels to showcase the sensitivity of the baseline algorithms. In addition, we propose two new observation patterns that are inhomogeneous in space-time: \textit{Sine} (a single sensor with a sine trajectory) and \textit{Exponential} (a full scan with exponentially distributed waiting times) (see Figure~\ref{fig:ks-obs}). The full details of the experimental setup are presented in Appendix~\ref{appendix-dataset}.

\begin{figure}
    \centering
    \includegraphics{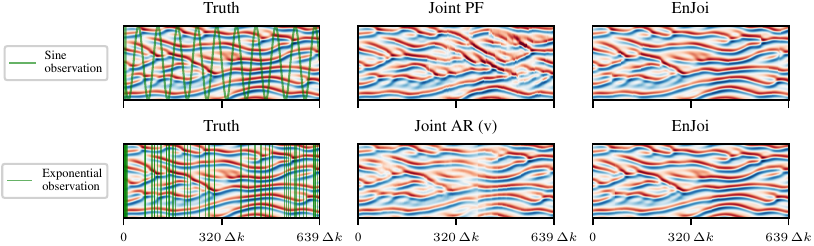}
    \caption{Kuramoto Sivashinsky assimilation experiment with non-homogneous observation patterns. The two left columns show the mean of an ensemble of 10 particles.}
    \label{fig:ks-obs}
\end{figure}
\paragraph{Results} 
Table \ref{tab:ks} presents the evaluation scores of the filtering algorithms on a dataset of 10 trajectories of 640 steps. The approaches that are too confident in the past state (e.g. the \textit{conditional} models) are failing, while the \textit{volatile} models perform better. The baselines are not equally robust to the different observation noise levels, in particular the volatile Joint AR fails in the high noise setting. The Joint PF obtains robust results in situations where the observations are homogeneous in space. However, it fails in the \textit{Sine} experiment with the moving sensor. We attribute this failure to its incapacity to correct the ensemble locally in space. Overall, our EnJoi algorithm is the most robust to all scenarios.

\begin{table}
    \centering
    \caption{Kuramoto Sivashinsky CRPS ($\downarrow$) average $\pm 3$ standard errors on the mean. We denote \textit{volatile} models with (v) and regular \textit{steady} models with (s), the guidance method is (g)}
    \label{tab:ks}
    \footnotesize
\begin{tabular}{lrrrrr}
\toprule
Obs. shape $H(k)$ & \multicolumn{3}{c}{Bernoulli} & \multicolumn{1}{c}{Sine} & \multicolumn{1}{c}{Exponential} \\
\cmidrule(lr){2-4} \cmidrule(lr){5-5} \cmidrule(lr){6-6}
Obs. noise std. $\sigma_{\text{obs}}$ & 0.1 & 0.5 & 2 & 0.5 & 0.5 \\
\midrule
Cond. P.F. (s) (g=MMPS) & 0.568 {\scriptsize $\pm$ 0.000} & 1.042 {\scriptsize $\pm$ 0.133} & 1.029 {\scriptsize $\pm$ 0.094} & 0.976 {\scriptsize $\pm$ 0.144} & 0.921 {\scriptsize $\pm$ 0.126} \\
Joint AR (s) (g=SDA) & 0.491 {\scriptsize $\pm$ 0.136} & 0.619 {\scriptsize $\pm$ 0.092} & 0.694 {\scriptsize $\pm$ 0.031} & 0.598 {\scriptsize $\pm$ 0.058} & 0.591 {\scriptsize $\pm$ 0.051} \\
Cond. AR (s) (g=SDA) & 0.457 {\scriptsize $\pm$ 0.096} & 0.606 {\scriptsize $\pm$ 0.128} & 0.576 {\scriptsize $\pm$ 0.087} & 0.574 {\scriptsize $\pm$ 0.086} & 0.571 {\scriptsize $\pm$ 0.078} \\
Joint AR (v) (g=scaled) & 0.088 {\scriptsize $\pm$ 0.010} & 0.093 {\scriptsize $\pm$ 0.008} & \textbf{0.200} {\scriptsize $\pm$ 0.017} & 0.516 {\scriptsize $\pm$ 0.036} & 0.213 {\scriptsize $\pm$ 0.026} \\
Joint P.F. (v) (g=scaled) & 0.013 {\scriptsize $\pm$ 0.001} & \underline{0.063} {\scriptsize $\pm$ 0.003} & 0.239 {\scriptsize $\pm$ 0.014} & 0.546 {\scriptsize $\pm$ 0.129} & \underline{0.120} {\scriptsize $\pm$ 0.019} \\
Joint AR (v) (g=SDA) & \underline{0.012} {\scriptsize $\pm$ 0.001} & 0.091 {\scriptsize $\pm$ 0.003} & 0.407 {\scriptsize $\pm$ 0.011} & \underline{0.293} {\scriptsize $\pm$ 0.049} & 0.136 {\scriptsize $\pm$ 0.017} \\
EnJoi (s) (ours) & \textbf{0.010} {\scriptsize $\pm$ 0.000} & \textbf{0.050} {\scriptsize $\pm$ 0.004} & \underline{0.229} {\scriptsize $\pm$ 0.054} & \textbf{0.162} {\scriptsize $\pm$ 0.069} & \textbf{0.097} {\scriptsize $\pm$ 0.029} \\
\bottomrule
\end{tabular}
\end{table}

\subsection{2D Kolmogorov Flow}

\begin{figure}[!htb]
    \centering
    \begin{minipage}[t]{0.48\textwidth}
        \centering
        \includegraphics[width=0.8\textwidth]{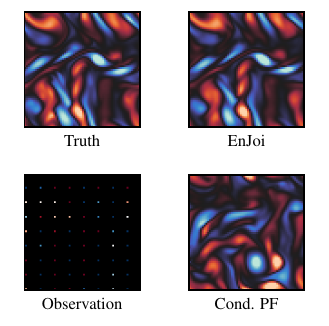}
        \caption{Final timestep of a Kolmogorov Flow assimilation experiment with sparse (8x8) observations. The left column displays one sample from the filter ensemble.}
        \label{fig:kolmogorov}
    \end{minipage}
    \hfill
    \begin{minipage}[t]{0.48\textwidth}
        \centering
        \includegraphics[width=\textwidth]{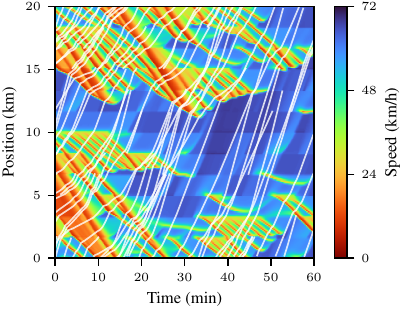}
        \caption{ARZ traffic flow simulation on a ring road with ramps. The velocity field is observed by only 2.5\% of vehicles, their trajectory on the domain is shown in white.}
        \label{fig:arz-sim}
    \end{minipage}
\end{figure}

\paragraph{Data} This experiment considers a 2D simulation of Navier Stokes flow with random forcing. The observations are laid out on a sparse grid (see Figure~\ref{fig:kolmogorov}) and we add a small observation noise $\sigma_{\text{obs}}=0.1$. The setup is similar to \cite{savaryTrainingFreeBayesianFiltering}, although our domain is half as wide (64x64). See Appendix D.2 in \cite{shysheyaConditionalDiffusionModels2024} for the data generation process.

\paragraph{Results} Table~\ref{tab:kolmogorov} shows averaged results over 10 trajectories of 64 steps. The conditional PF obtains a better performance than i.i.d.\ runs of Conditional AR in the dense observation regime (16x16), demonstrating the usefulness of particle resampling. However the results do not hold for sparse observations and the Joint models reach better CRPS. In the denser regimes, the performance of Joint AR and EnJoi is comparable, which is expected because of the low noise and static sensor positions.
\begin{table}
    \centering
    \footnotesize
    \caption{Kolmogorov CRPS ($\downarrow$) average $\pm 3$ standard errors on the mean.}
    \label{tab:kolmogorov}
    \begin{tabular}{lrrr}
\toprule
Algorithm & \multicolumn{3}{c}{Observed grid points (out of 64x64)} \\
\cmidrule(lr){2-4}
 & $(4, 4)$ & $(8, 8)$ & $(16, 16)$ \\
\midrule
Joint AR (v) (g=scaled) & 0.31 {\scriptsize $\pm$ 0.04} & 0.247 {\scriptsize $\pm$ 0.028} & 0.083 {\scriptsize $\pm$ 0.007} \\
Cond. P.F. (s) (g=MMPS) & 0.32 {\scriptsize $\pm$ 0.04} & 0.245 {\scriptsize $\pm$ 0.082} & 0.043 {\scriptsize $\pm$ 0.009} \\
Cond. AR (s) (g=SDA) & 0.23 {\scriptsize $\pm$ 0.03} & 0.186 {\scriptsize $\pm$ 0.024} & 0.074 {\scriptsize $\pm$ 0.018} \\
Joint AR (v) (g=SDA) & \underline{0.18} {\scriptsize $\pm$ 0.02} & \textbf{0.043} {\scriptsize $\pm$ 0.002} & \underline{0.036} {\scriptsize $\pm$ 0.001} \\
EnJoi (s) (ours) & \textbf{0.12} {\scriptsize $\pm$ 0.02} & \underline{0.047} {\scriptsize $\pm$ 0.002} & \textbf{0.029} {\scriptsize $\pm$ 0.001} \\
\bottomrule
\end{tabular}
\end{table}

\subsection{Highway traffic PDE}

\paragraph{Data}
We use a macroscopic second-order traffic simulation based on the PDE from \citet{awResurrectionSecondOrder2000,zhangNonequilibriumTrafficModel2002}. The domain is set to a 20km-long ring road with 12 equally spaced ramps. The exit-entry ratios are simulated with random step functions and act as perturbations. We simulate 1 hour of traffic using a finite-volume solver (more details in Appendix~\ref{appendix-dataset}). The observations consist of simulated trajectories from a small portion of probe vehicles that move over the domain, following the mean velocity with a probe-specific variance (See Figure \ref{fig:arz-sim})

\begin{wraptable}{r}{0.5\textwidth}
% \vspace{-\baselineskip}
\centering
\caption{ARZ traffic CRPS ($\downarrow$) average $\pm$ 3 standard errors on the mean}
\label{tab:traffic}
\footnotesize
\begin{tabular}{lrr}
\toprule
Algorithm & \multicolumn{2}{c}{Penetration rate} \\
\cmidrule(lr){2-3}
 & 2.5\% & 5\% \\
\midrule
Joint AR (s) (g=scaled) & \underline{0.21} {\scriptsize $\pm$ 0.14} & 0.21 {\scriptsize $\pm$ 0.14} \\
Joint P.F. (s) (g=MMPS) & 0.23 {\scriptsize $\pm$ 0.05} & \underline{0.18} {\scriptsize $\pm$ 0.05} \\
EnJoi (s) (ours) & \textbf{0.15} {\scriptsize $\pm$ 0.06} & \textbf{0.11} {\scriptsize $\pm$ 0.03} \\
\bottomrule
\end{tabular}
% \vspace{-\baselineskip}
\end{wraptable}

\paragraph{Results} This experiment highlights the heterogeneity of real world measurements. With moving sensors, all the sections of the highway are not equally observed, creating a moving uncertainty pattern. The traffic PDE exhibits shockwaves and rapid shift of conditions, making it challenging for algorithms with high inertia. To demonstrate that local correction is more effective than resampling, we use 128 particles for the particle filter and 10 for the other methods. Our algorithm obtains the best reconstruction results, see Table~\ref{tab:traffic}.

\section{Conclusion}

In this work we have argued that current generative data assimilation algorithms are limited by their strict auto-regressive conditioning. We proposed to take advantage of a flexible \textit{joint} generative model which we condition on a more expressive distribution of background states. Our new filtering algorithm is able to better balance past and future uncertainty and outperforms all baselines on simulated experiments. Its robustness to non-homogeneous observation patterns makes it applicable to a wide range of domains.

One limitation of our method is the increased computation cost: joint scores almost double the inference time and the covariance inverse also adds overhead (see Appendix~\ref{appendix-times}). Another limit is that scaling the guidance with the empirical covariance degrades the rollout quality (see the forecast in Figure~\ref{fig:ks-combined}) and necessitates Langevin corrections. Similarly to En4DVar, we make the simplifying assumption that the background is a Multivariate Gaussian, thus we are not able to handle multimodal past priors. An interesting future direction would be to consider more expressive background priors, using tools from Stochastic Interpolants or Optimal Transport.

\subsection*{AI use statement}

% (This section is \textbf{required} and does not count toward the page limit.)

In this work, we used generative AI tools to implement methods.
We have not used generative AI tools to generate synthetic data sets, help develop theoretical models or conceptual frameworks, assist with translation, clean and reformat dataset, support qualitative and thematic data analysis, interpret results and design or provide feedback on research methodology or experiments. The following are not applicable to this work: formulate mathematical claims, provide critical ingredients for proving mathematical claims, assist in the writing of proofs, propose or refine hypotheses.
Additionally, we used generative AI tools to create or modify scientific figures or images, create or edit software code, summarize or analyse existing literature, discover research topics or identify gaps, brainstorming, sourcing/searching for information, edit a research paper to improve readability, identify relevant literature. We have reviewed all AI-assisted work. LLM-generated code was verified and tested for correctness. We take responsibility for the final content of this work, including text, claims or artifacts produced with the aid of generative AI.

% See the ICLR 2027 AI Policy for Authors for more details. This statement should
% not be more than 1 page.

\subsection*{Ethics statement}

% (This section is \textbf{recommended} and does not count toward the page limit.)
This paper uses generative AI for the prediction of dynamical systems. Careful evaluation of the model bias and robustness should be conducted before using these predictions for real-world decision making.
% If authors feel that their paper submission raises questions regarding the Code
% of Ethics, they are encouraged to include a paragraph of Ethics Statement (at
% the end of the main text before references) to address potential concerns where
% appropriate. Topics include, but are not limited to, studies that involve human
% subjects, practices to data set releases, potentially harmful insights,
% methodologies and applications, potential conflicts of interest and sponsorship,
% discrimination/bias/fairness concerns, privacy and security issues, legal
% compliance, and research integrity issues (e.g., IRB, documentation, research
% ethics). This statement should not be more than 1 page.

\subsection*{Reproducibility statement}

The code necessary for reproducing experiments of Section~\ref{section-experiments} is available at \href{https://github.com/J-Moreau/enjoi}{https://github.com/J-Moreau/enjoi}.
Details on how to generate the synthetic datasets are in Appendix~\ref{appendix-dataset}.
% (This section is \textbf{recommended} and does not count toward the page limit.)

% It is important that the work published in ICLR is reproducible. Authors are
% strongly encouraged to include a paragraph-long Reproducibility Statement at the
% end of the main text (before references) to discuss the efforts that have been
% made to ensure reproducibility. This paragraph should not itself describe
% details needed for reproducing the results, but rather reference the parts of
% the main paper, appendix, and supplemental materials that will help with
% reproducibility. For example, for novel models or algorithms, a link to an
% anonymous downloadable source code can be submitted as supplementary materials;
% for theoretical results, clear explanations of any assumptions and a complete
% proof of the claims can be included in the appendix; for any datasets used in
% the experiments, a complete description of the data processing steps can be
% provided in the supplementary materials. Each of the above are examples of
% things that can be referenced in the reproducibility statement.



\bibliography{references}
\bibliographystyle{iclr2027_conference}

\appendix

\section{Details on Diffusion Models}
\label{appendix-diffusion}

\subsection{Score-based Generative Models}

Denoising Diffusion models \citep{hoDenoisingDiffusionProbabilistic2020,songScoreBasedGenerativeModeling2021} generate samples $\xo \in \mathbb{R}^d$ of a target data distribution $p_{\text {data}}(\xo)$ starting from initial Gaussian noise $\x_1 \sim \mathcal{N}(\x_1;0,I)$. Let $t \in [0,1]$, we define $a_t,b_t \in \mathbb{R}$ the scale and noise coefficients and $\x_t$ a random variable following the Gaussian probability path $\x_t=a_t\x_0 + b_t\x_1$. At all times, we have the relation $\x_t|\x_0 \sim \mathcal{N}(\x_t\;|\;a_t \x_{0}, b_t^2 I)$. From these coefficients, we can define the forward diffusion SDE \eqref{eq:forward} which noises the input data until it becomes pure Gaussian noise. The associated family of backward SDEs \eqref{eq:backward}  generates clean samples from noise, for any  $\eta \in \mathbb{R}$.
\begin{align}
    \mathrm{d}\overrightarrow{\x_t} &= f_t \x_t\,\mathrm{d}t + g_t\,\mathrm{d}\mathbf{w}_t \label{eq:forward}\\
    \mathrm{d}\overleftarrow{\x_t} &= \left[f_t \x_t - \frac{1+\eta^2}{2}g_t^2 \nabla_{x_t} \log p(\x_t)\right]\mathrm{d}t + \eta g_t\,\mathrm{d}\mathbf{w}_t \label{eq:backward}
\end{align}
Where $w_t$ is the standard Wiener process in $\mathbb{R}^d$, $f_t\in \mathbb{R}$ and $g_t \in \mathbb{R}^+$ are derived from $a_t$ and $b_t$. In this paper we use the Variance Preserving SDE with coefficients $a_t = \sqrt{ \alpha_t }$ and $b_t = \sqrt{1 -\alpha_t}$, $\alpha_t \in \mathbb{R}^+$, which grants the property $a_t^2 + b_t^2=1$. The score $\nabla_{x_t}\log p (x_t)$ is approximated with a Neural Network $s_\theta(x_t,t)$ trained by minimizing the score-matching loss \eqref{eq:score-matching}. In practice, it is more stable to parametrize the score network as a function of the noise-predictor $\epsilon_\theta(x_t,t) = -b_t s_\theta(x_t,t)$, which shares the same minimizer.
\begin{align}
    \mathcal{L}(\theta) = 
    \mathbb{E}_{\substack{t,\,x_0,\,x_1}}
    \left[\left\|\epsilon_\theta(a_t x_0+b_t x_1,t)-x_1\right\|_2^2\right]
    \label{eq:score-matching}
\end{align}

Once the score has been trained, there exists a variety of sampling methods to numerically integrate Equation~\eqref{eq:backward} \citep{songDenoisingDiffusionImplicit2022, karrasElucidatingDesignSpace2022,luDPMSolverFastSolver2025}. Following \citet{shysheyaConditionalDiffusionModels2024}, we use the DPM solver \cite{luDPMSolverFastODE2022}.

\subsection{Diffusion Posterior Sampling}
\label{appendix-dps}
Guidance is a test-time method to sample from $p(\x|\y)$ with $\y$ not seen during training. Assuming that we have access to a likelihood function $p(\y|\xo)$, we can condition the score on observations by adding a guidance term:
\begin{align*}
p(\xt|\y) &= p(\xt)p(\y|\xt)/p(\y) \\
\nabla_{\xt} \log p(\xt|\y) &= \nabla_{\xt} \log p(\xt) + \nabla_{\xt} \log p(\y|\xt)
\end{align*}
The last term is intractable and needs to be approximated. DPS \citep{chungDiffusionPosteriorSampling2024} proposes to substitute $\xt$ with the denoising expectation obtained with Tweedie's formula $\tildexo(\xt)= \mathbb{E}[\xo|\xt]=\frac{1}{a_t}(\xt + b_t^2 s_\theta(\xt,t))$. It is common to model $p(\xo|\xt)$ as a Gaussian to account for the denoising error.
\begin{equation}
    p(\y|\xt) \approx\int{p(\y|\xo)\mathcal{N}(\xo|\tildexo(\xt),\mathbf{\tilde{\Sigma}}(\xt))\,\mathrm{d}\xo} \label{eq:dps-approx}
\end{equation}
 Where $\mathbf{\tilde{\Sigma}}(\xt)$ is an approximation of the variance $\mathbb{V}[\xo|\xt]$. SDA \citep{rozetScorebasedDataAssimilation2023} uses $\mathbf{\tilde{\Sigma}}(\xt) = (b^2_t/a_t^2) \,\mathbf{\Gamma}$, where $\mathbf{\Gamma}$ is a constant.
% For a better estimator, \citet{rozetLearningDiffusionPriors2024} propose Moment-matching Posterior Sampling (MMPS) which uses the explicit Tweedie covariance formula $\mathbf{\tilde{\Sigma}}(\xt) = b^2_t \nabla_{\xt}\mathbb{E}[\xo|\xt]$.
% The method proposed by \cite{rozetLearningDiffusionPriors2024} uses conjugate gradients instead of inverting the matrix. The practical algorithm computes jacobian vector products with automatic differentiation to avoid materializing the matrix.
In the case of Gaussian observation error $p(\y|\xo) = \mathcal N(\y|\mathbf{H}\xo,\mathbf{R})$, the guided score is
\begin{align}
    \nabla_{\xt} \log p(\xt|\y) &\approx s_\theta(\xt,t) - \frac{1}{2}\nabla_{\xt}\|\y-\mathbf{H}\tildexo(\xt)\|^2_{\mathbf{\tilde{R}}} \quad \text{where }
    \mathbf{\tilde R} = \mathbf{R}+\mathbf{H}\mathbf{\tilde{\Sigma}}\mathbf{H}^T
    \label{eq:dps2}
\end{align}
The gradient is computed by auto-differentiating through the denoiser. The approximation \eqref{eq:dps-approx} and the sampler discretization often cause the state to deviate from the exact denoising manifold induced by $p(\xt)$. This issue is exacerbated when the guidance signal is large, thus a solution is to stabilize the sampling with Langevin corrections (see Equation~\eqref{eq:langevin})
\begin{align}
    \xt &\gets \xt + \tau \cdot \frac{\nabla_{\xt}\log p(\xt|\y)}{\|\nabla_{\xt} \log p(\xt|\y)\|} + \sqrt{\tau} \mathbf{\epsilon_c}, \quad \epsilon_c \sim \mathcal{N}(\mathbf{0},\mathbf{I})
    \label{eq:langevin2}
\end{align}

\section{Related works}
\label{appendix-scores}

\subsection{Trajectory sampling}

This section provides additional details on the related sequential diffusion sampling approaches. Also see Table~\ref{tab:trajectory-sampling} for comparison.

\begin{table}[tb]
    \centering
    \caption{Comparison between diffusion trajectory sampling methods.}
    \label{tab:trajectory-sampling}
    \footnotesize
    \begin{tabularx}{\linewidth}{@{}lllll@{}}
        \toprule
        &Learned score & Observation guidance & Past guidance\\
        Sampling Method $\quad \nabla \log p =$ & $s_\theta(.,t)$
        & $+\frac{1}{2}\nabla\|\y-H\tildexo\|^2_R$
        & $+\frac{1}{2}\nabla\|\x^k-\tildexko\|^2_P$ \\
        \midrule
        \citet{rozetScorebasedDataAssimilation2023} & joint & on all timesteps & no \\
        \citet{savaryTrainingFreeBayesianFiltering} & conditional & on $k+1$ & no \\
        \citet{shysheyaConditionalDiffusionModels2024} & joint & on $k+1$ & constant $\mathbf{P}$ \\
        Ours & joint & on $k+1$ & empirical $\mathbf{P}^k$ of particles \\
        \bottomrule
    \end{tabularx}
\end{table}

\paragraph{SDA}
\citet{rozetScorebasedDataAssimilation2023} introduce the smoothing score:
\begin{align*}
    p(\xkkpt|\ykkp)&\propto p(\xkkpt)p(\ykkp|\xkkpt)\\
        & \underset{\text{DPS}}{\approx} p(\xkkpt)\mathcal{N}(\ykkp|H\tildexkkpo,\tilde{R})\\
    \nabla_{\xkkpt} \log p(\xkkpt|\ykkp) &\approx \underbrace{s_\theta(\xkkpt)}_{\text{joint score}} -\underbrace{ \frac{1}{2} \nabla_{\xkkpt}\|\ykkp-H\tildexkkpo(\xkkpt)\|^2_{\tilde R} }_{\text{guidance on all observations}}
\end{align*}
Notice that the probability only depends on observations $\ykkp$ not $\yonekp$, thus there is no background (past) information. This can be extended to any horizon $k:k+n$, and is meant to be used on the full window $0:T$ with observations at all timesteps. The algorithm isn't meant for online filtering, or simulation rollout.

\paragraph{Joint AR} To include the background state into an auto-regressive scheme, \citet{shysheyaConditionalDiffusionModels2024} combines the joint score with reconstruction guidance \citep{hoVideoDiffusionModels2022}.
\begin{align}
    p(\xkkpt|\xk,\ykp)&\propto p(\xkkpt)p(\xk,\ykp|\xkkpt) \notag \\
            &\approx p(\xkkpt)\overbrace{p(\xk|\xkkpt)p(\ykp|\xkkpt)}^{\approx \text{independent}} \notag\\
            & \underset{\text{DPS}}{\approx} p(\xkkpt)\mathcal{N}(\xk|\tildexko,\tilde{\Gamma})\mathcal{N}(\ykp|H\tildexkpo,\tilde{R})\notag\\
        \nabla_{\xkkpt} \log p(\xkkpt|\xk,\ykp) &\approx \underbrace{s_\theta(\xkkpt)}_{\text{joint score}} - \underbrace{\frac{1}{2}\nabla_{\xkkpt} \|\xk - \tildexko\|^2_{\tilde\Gamma}}_{\text{auto-regressive guidance}} \notag \\ &\quad - \underbrace{\frac{1}{2}\nabla_{\xkkpt} \|\ykp-H\tildexkpo\|^2_{\tilde{R}}}_{\text{guidance on observation }k+1}  \label{eq:rec-sda}
\end{align}

With $\tilde\Gamma=\mu_t^2/\sigma_t^2 \Gamma + \zeta$ where $\zeta$ is a small value. Because the guidance signal needs to be strong enough to condition the generation, sampling needs to be stabilized with Langevin corrections.

\paragraph{Conditional AR}
\citet{shysheyaConditionalDiffusionModels2024} and \citet{savaryTrainingFreeDataAssimilation2025} make use of an \textit{amortized} score - i.e. conditioned by architecture. The input of the score network is a clean past state $\xk$ and a noisy future state $\xkpt$.
\begin{align}
    p(\xkpt|\xk,\ykp)&\propto p(\xkpt|\xk)p(\ykp|\xkpt,\xk)\\
        & \underset{\text{DPS}}{\approx} p(\xkpt|\xk)\mathcal{N}(\ykp|H\tildexkpo,\tilde{R})\\
    \nabla_{\xkpt} \log p(\xkpt|\xk,\ykp) &\approx \underbrace{s_\theta(\xkpt|\xk)}_{\text{conditional score}} - \underbrace{\frac{1}{2}\nabla_{\xkpt} \|\ykp-H\tildexkpo(\xkpt|\xk)\|^2_{\tilde{R}}}_{\text{guidance on observation }k+1} \label{eq:cond-sda2}
\end{align}
This is ideal for building an Autoregressive diffusion, where we fully trust the previously generated sample $\xk$. \citet{shysheyaConditionalDiffusionModels2024} report superior performance on forecasting tasks, compared to a model that is conditioned via guidance \ref{eq:rec-sda}. Conditioning via architecture ensures stabler rollouts, there is no need to use Langevin corrections since the general direction is already well determined. However, it cannot easily deviate from its established course.

\paragraph{Stochastic Interpolants} FlowDAS \citep{chenFlowDASStochasticInterpolantbased2025} uses a variant of conditional SDA for stochastic interpolants, where the model learns a path from $\xk$ to $\xkp$. The follow-up \textbf{ForcingDAS} \citep{jiaForcingDASUnifiedRobust2026} combines FlowDAS to Diffusion Forcing \citep{chenDiffusionForcingNexttoken2024} to handle arbitrary noise levels. The filter "re-noises" the past states, enabling the background to be progressively forgotten. However the noise schedule is based on a fixed heuristic and does not adapt to the model uncertainty.

\subsection{Ensemble of trajectories}
\label{appendix-related-works}
In the past section, we have reviewed sampling algorithms for generating a single trajectory. For completeness, we now present the result of recent research on \textit{ensemble} filters that operate with a distribution of trajectories.
% \paragraph{CDSB} Schrödinger Bridges (Shi et al. 2022) TODO read this
\textbf{FA-APF} \citep{savaryTrainingFreeBayesianFiltering} is a particle filter that uses the optimal proposal from a conditioned diffusion model \eqref{eq:cond-sda}. This is a practical algorithm that works out-of-the-box with a diffusion model that has already been trained on massive data (e.g. GenCast). However, this framework doesn't allow for much deviation from the past state, hence particles with incorrect background are eliminated. \textbf{SURGE} \citep{weiSURGEApproximationfreeTraining2026} is a particle filter variant that uses a stochastic interpolant to sample $p(\xkp|\xk)$. The novelty is to use particle weighting and resampling at each step of the SDE integration which results in better observation guidance. These particle filter methods suffer from the high-dimensionality curse and localization issues. There is a low probability that a particle background will be correct in all dimensions, but because particle likelihoods are computed globally, particles that are only locally correct are eliminated.

\subsection{Numerical Solver hybrids}
Another research direction is to assume that we have access to a numerical forward model of the dynamics at inference time. The following methods use a generative framework in conjonction with a simulator.

% The model learns to sample from the unconditional $p(\x)$ with a SDE from $\mathcal{N}(0,I)$ to $\x$. The conditioning comes from guidance on $\ykp$ and initializing the SDE from a noisy $\epsilon_{\text{init}}$. The initial noise is obtained by integrating the SDE in reverse, starting from the forwarded state $\xkp= \mathcal{A}(\xk)$ to an intermediate noise level. This achieves background and observation mixing, although there is no principled way to tune the initial noise level. The particles are essentially i.i.d.\ so the background doesn't benefit from other particle information.

\paragraph{Non-Deep Models} \textbf{EnSF} \citep{baoNonlinearEnsembleFiltering2025} is a Particle Filter that approximates the diffusion score using the numerical simulator instead of a neural netwok. This Monte Carlo approach is easier to implement but we believe that using a learned physical model is more efficient in terms of ensemble size.

\paragraph{Latent Data Assimilation} Prior work has been conducted on hybridation of Assimilation algorithms with deep latent states. Notably using Variational Autoencoders to transpose the problem to a low-dimensional space with Gaussian distribution \citep{_zalp_2026}. These reduced-order methods are less compute-intensive, however diffusion-based approaches have shown better generative performance. Thus, a solution is to combine latent-space encoding with score-based sampling \cite{rozetLostLatentSpace, si2025latent}. In the same vein, \textbf{DAISI} \citep{andraeDAISIDataAssimilation2026} projects the prediction of a simulator to a noisy latent space, then uses a guided denoiser to assimilate measurements. Adding noise to the latent reduces the trust in the background, but there is no principled method to determine the correct level.

\section{Limitations of the joint formulation}

\subsection{Background prior}
\label{appendix-prior-swap}
In Bayesian Filtering, there is a fundemental difference between the observation and the background information: the measurements provide a likelihood while the past acts a prior. This is because the model of the dynamics is usually given as a conditional probability $p(\xkp|\xk)$. This is seen in the classical 4DVar decomposition:
\begin{equation}
    p(\xk,\xkp|\yonekp) = \underbrace{p(\ykp|\xkp)}_{\text{obs. likelihood}}\;\;\underbrace{p(\xkp|\xk)}_{\text{conditional model}}\;\; \underbrace{\mathcal{N}(\xk|\hatxk,\mathbf{P}^k)}_{\text{background prior}} \label{eq:conditional-prob}
\end{equation}
In our sampling algorithm, we choose to see the background as a likelihood.
\begin{equation}
    p(\xk,\xkp|\yonekp) \approx \underbrace{p(\ykp|\xkp)}_{\text{obs. likelihood}}\;\;\underbrace{p(\xk,\xkp)}_{\text{joint model}}\;\; \underbrace{\mathcal{N}(\xk|\hatxk,\mathbf{P}^k)}_{\text{background likelihood}} \label{eq:joint-prob}
\end{equation}
We have chosen this approximation because it is straight-forward for a diffusion model to learn the joint noised score $\nabla \log p(\xkt,\xkpt)$. On the other hand, learning the score $\nabla \log p(\xkpt|\xkt)$ with a noisy conditional $\xkt$ is non-standard (see the next section). Using a conditional score with clean past $\nabla \log p(\xkpt|\xk)$ is also straight-forward but doesn't offer the same flexibility.

Because of this choise, our proposed sampling algorithm has a bias, even in the linear Gaussian setting. It uses the model $p(\xk,\xkp)$ to generate $\xk$ rather than sampling from the background Gaussian prior $\mathcal N(\xk|\hatxk,\Pk)$. As a result, it will be biased towards the static (i.e.\ climatological) distribution rather than trusting the background trajectory. However, because the Gaussian background assumption is physically inacurrate, being biased towards the learned model could instead be beneficial to our algorithm.

One hypothetical drawback is for trajectories that fall in the tails of the training distribution. In this situation, we believe that the state will eventually return to a domain that it has seen more frequently in training. For completeness, we describe an alternative method that avoids the prior-likelihood swap. The next section describes a more expensive and arguably less-biased version of our method that could be explored in future work.

\subsection{Noisy Conditional score}

In an attempt to apply 4DVar to a diffusion model, we have proposed to use the joint formulation \eqref{eq:joint-prob}. However, it is more natural to try and generalize Equation~\eqref{eq:conditional-prob} to the noisy setting. An alternative to the \textit{joint} denoising of Equation~\eqref{eq:two-guidances} is to perform a \textit{conditional} denoising $\xkpt|\xkt$.
\begin{align}
    p(\xkkpt|\yonekp) \approx \underbrace{p(\ykp|\tildexkpo)}_{\text{obs. likelihood}}\;\;\underbrace{p(\xkpt|\xkt)}_{\text{conditional model}}\;\; \underbrace{\mathcal{N}(\tildexko|\hatxk,\mathbf{P}^k)}_{\text{background prior}} \label{eq:noisy-conditional-prob}
\end{align}

In the sampling algorithm, we will need the score $\nabla \log p(\xkpt|\xkt)$, where the gradient is taken with respect to $\xkt$ and $\xkpt$. Fortunately, the conditional score can be expressed as a sum of two denoising-diffusion score networks:
\begin{align}
    \nabla_{\xkt}\log p(\xkpt|\xkt) &= \underbrace{\nabla_{\xkt}\log p(\xkt,\xkpt)}_{\text{left part of joint score}} - \underbrace{\nabla_{\xkt}\log p(\xkt)}_{\text{marginal score}}\\
    \nabla_{\xkpt}\log p(\xkpt|\xkt) &= \underbrace{\nabla_{\xkpt}\log p(\xkt,\xkpt)}_{\text{right part of joint score}}
\end{align}
The marginal score $\nabla_{\xkt}\log p(\xkt)$ can be approximated by a different neural-network, using the classical denoising-score matching objective \eqref{eq:score-matching}. Essentially, the marginal score network erases the climatological prior on $\xk$, so it can be replaced by the background prior.

This opens the way for a generalization of the current method, although we have not verified how it performs in practice. The marginal score requires a second neural network in both training and inference which creates a computation overhead.

\section{Experiment details}

\label{appendix-dataset}
\subsection{Kuramoto-Sivashinsky}
The dataset consists in synthetic 1D fluid data generated with the following PDE:
\begin{align}
    \frac{\partial u}{\partial k} + u\frac{\partial u}{\partial z} + \frac{\partial^2 u}{\partial z^2} + \frac{\partial^4 u}{\partial z^4} = 0
\end{align}
The spatial axis is decomposed into 256 cells $\{i \Delta z\}_{i=0}^L$ of length $\Delta z= 1/4$, the total time is 120 seconds which is split into 640 sub-timesteps $\Delta k$. The joint score network is trained on temporal windows of width $9\Delta k$. For conditional scores and inference, we decompose the window into two timesteps: conditioning $k=\{0,1,2,3\}$ and prediction $k+1=\{4,5,6,7,8\}$.

The Neural Networks are 1D UNet with 12 hidden layers, we use the base implementation of the "SDA" architecture from \cite{shysheyaConditionalDiffusionModels2024} which we train for 1000 epochs of 4096 windows on an Nvidia T4 GPU. In their implementation, the temporal steps are simply represented as channels of the discretized field. The conditional model uses the same architecture as the Joint and learns to decode both past and future, but is fed the past state as an extra input.

The observations are sampled from $\mathbf{y}= \mathbf{H}\mathbf{u} + \boldsymbol{\rho}$ where $\mathbf{H}$ is a boolean mask and $\boldsymbol{\rho} \sim \mathcal{N}(\mathbf{0}_Y,r\mathbf{I}_Y)$ is i.i.d noise. We experiment with three observation shapes:
\begin{align}
    \forall i,j &\quad \mathbf{H}[i,j] \sim \mathcal{B}(p) &\quad \text{Sparse Bernoulli} \label{eq:bernoulli} \\
    \forall j,\Delta j &\quad p(\mathbf{H}[:,j+\Delta j]\;|\;\mathbf{H}[:,j]) = \mathcal{E}(\lambda\Delta j) &\quad \text{Exponential waiting time} \label{eq:exponential}\\
    \forall i,j &\quad  \mathbf{H}[i,j] = 1 \; \text{ if }\; i \Delta z \approx L/2 + L\sin(\omega j) &\quad \text{Oscillating observer} \label{eq:sine}
\end{align}
The Bernoulli probability is set to $p=0.1$, the waiting time is $\lambda=10$ and the frequency is $\omega=1/64$.

\subsection{Kolmogorov Flow}

The experiment uses a Navier-Stokes simulation on a periodic 2D domain. The velocity field $\mathbf{u}$ is described by the following system:
\begin{align*}
    \partial_k \mathbf{u}+ \mathbf{u} \cdot \Delta \mathbf{u} - \frac{1}{\operatorname{Re}} \Delta^2 \mathbf{u} + \frac{1}{\rho}\Delta p  = \mathbf{f}\\
    \nabla \cdot \mathbf{u} = 0
\end{align*}
Where $p$ is the pressure field, $\operatorname{Re}=1000$ is the Reynolds number, $\rho=1$ is the density and $\mathbf{f}$ is a forcing term. We refer to the paper from \citet{rozetScorebasedDataAssimilation2023} for details on the generation process. The discrete velocity field is of size (2x64x64), the total duration of the trajectories is 64 timesteps.

We train the models on windows of length 5 with $k=[\![1,2]\!]$ and $k+1=[\![3,4,5]\!]$.
The Neural Network is a 2D UNet with 24 hidden layers, we use the base implementation of the "SDA" architecture from \cite{shysheyaConditionalDiffusionModels2024} which we train for 1000 epochs of 4096 windows on an Nvidia A100 GPU.

\subsection{Road Traffic}
The ARZ equations are a classic macroscopic model of motorway traffic on a one-dimensional road $z \in [0,L]$. The PDE system describes the evolution of vehicle density $\rho(z,k)$ (vehicles/meter) and average velocity $v(z,k)$. The first equation imposes the conservation of vehicles, the second equation couples the density $\rho$ with the velocity $v$ using a traffic pressure $h(\rho)$. 
\begin{align}
    \label{eq:arz}
\begin{cases}
    \partial_k \rho + \partial_z (v\rho) = 0 \\
    \partial_k (v + h(\rho)) + v \;\partial_z (v+h(\rho)) = \frac{1}{T_{\text{relax}}}(v - V_{\text{eq}}(\rho))
\end{cases}
\end{align}
We include 12 equally-spaced ramps to the simulated ring road. The ramps allow a fraction ($0.8 \leq \beta_r \leq 1.2$) of the traffic flow to enter or exit the road, this acts as an unpredictable exterior perturbation. We use the pressure term $h(\rho)$ and non-concave flux function $\rho V_{\text{eq}}(\rho)$ from David Ketcheson's codebase (\href{https://github.com/ketch/aw-rascle-zhang}{https://github.com/ketch/aw-rascle-zhang}) which we find to generate more stop-and-go waves than a classical Greenshields flux in our experiments.

The simulation domain is divided in a grid with time discretization $\Delta \kappa=0.1$s\ and cell length $\Delta z=50$m. We use the setup described in \citet{moreau4DVarAssimilationVehicle2026}, the numerical solver is a finite-volume Rusanov scheme. The intial state and ramp entry-exit ratios are generated using a random combination of step functions. The data is then downscaled to $\Delta k=10$s and $\Delta z=100$m and the velocity is normalized to $[-1,1]$ for learning and inference. The discretized state of the velocity field is of shape $(200)$ and the density field is not seen by the model.

We generate 600 trajectories of traffic flow as training data, 200 for validation and 200 for testing. We use a window of 12 time steps, $k=[\![ 1, 6]\!]$ for conditioning and $k+1=[\![ 7, 12]\!]$ for prediction. Although the observation pattern is not strictly linear and uncorrelated, the guidance is approximated as i.\ i.\ d.\ Gaussian taking the empirical observation RMSE at around $0.05$. The Neural Network is the same architecture as the KS dataset and we train for 1000 epochs of 4096 windows on an Nvidia A100 GPU. Because of the extended compute time, we only evaluate the models on 10 full simulations from the test dataset.

\subsection{Metrics}
\label{appendix-metrics}

We now present the metrics used in this study. We denote the predicted state $\x$ and ground truth $\x^\star$, both vectors contain space and time dimensions.
\paragraph{Skill} This is a direct measure of the prediction performance, it consists in the RMSE of the mean of the ensemble.
\begin{align}
    \operatorname{Skill} = \sqrt{\operatorname{Mean}\left(
    (\x^\star - \frac{1}{N}\sum_{i=1}^{N}\x^{(i)})^2\right)}
\end{align}
Where the Mean is taken with respect to the spatial and temporal dimension to obtain a scalar average.

\paragraph{Spread} This indicator of the sample variety is obtained by taking the  standard deviation of the ensemble members, which is then averaged over the temporal and state dimension.
\begin{align}
    \operatorname{Spread} = \sqrt{\operatorname{Mean}\left(\frac{1}{N}\sum_{i=1}^{N}
    (\x^{(i)}-\frac{1}{N}\sum_{j=1}^{N}\mathbf{x}^{(j)})^2\right)}
\end{align}
We use the biased standard deviation in our calculations.

\paragraph{CRPS} The Continuous Ranked Probability Score measures the error between the distribution derived from the prediction ensemble $\x^{(i)}, 1\leq i\leq N$ and the true trajectory $\x$. The result is then averaged over the temporal and state dimension.
\begin{align}
    \operatorname{CRPS} = \operatorname{Mean}\left( \frac{1}{N}\sum_{i=1}^{N}\left\|\x^{(i)}-\x^\star\right\|_1
    - \frac{1}{2N^2}\sum_{i=1}^{N}\sum_{j=1}^{N}\left\|\xki-\mathbf{x}^{k,(j)}\right\|_1 \right)
\end{align}

\subsection{Implementation details}

\paragraph{Sampling} We use the DPM solver \citep{luDPMSolverFastODE2022} with 128 steps for all methods. We use one Langevin step \eqref{eq:langevin} for every method except when using MMPS, the stepsize is chosen as $\tau=0.25$, the guidance strength for SDA is $\boldsymbol{\Gamma}=0.1\mathbf{I}$. The \textit{volatile} models (small R) use a modified observation guidance $\mathbf{R}=\sigma_{\min}^2 \mathbf{I}$ with $\sigma_{\min}=0.01$. In Figure~\ref{fig:ks-combined}, the model with (large P) has a modified reconstruction guidance $\mathbf{P}=\sigma_{\text{large}}^2 \mathbf{I}$ with $\sigma_{\text{large}}=0.5$.

\paragraph{Covariance localization} For all experiments, we use a spatial bandwidth of $3\Delta z$. The temporal bandwidth is chosen as $1\Delta k$ for Kolmogorov, $5\Delta k$ for ARZ and $10\Delta k$ for KS.

\paragraph{Joint PF} When computing likelihood weights in the Joint Particle Filter, we can't easily estimate the expected state $\mathbb{E}[\xkp|\xk]$ like in the original Conditional version. Thus we sample a prediction $\xkp \sim p(\xkp|\xk)$ for each particle and compute the likelihood of observations $\ykp$ given the generated $\xkp$.

\section{Compute times}

\label{appendix-times}

Table~\ref{tab:compute-times} reports the wall-clock time to assimilate a single trajectory for each method, averaged over 10 runs. 

\begin{table}[htb]
\centering
\caption{Assimilation time per single trajectory (h:mm). The KS and Kolmogorov experiments are run on an Nvidia P4 GPU and the ARZ experiment on an Nvidia T4 GPU. All methods use 10 particles, except the ARZ particle filter which uses 128.}
\label{tab:compute-times}
\footnotesize
\begin{tabular}{lrrrr}
\toprule
 & \multicolumn{2}{c}{KS} & Kolmogorov & ARZ \\
\cmidrule(lr){2-3} \cmidrule(lr){4-4} \cmidrule(lr){5-5}
 & P4 & T4 & P4 & T4 \\
Method & & & & \\
\midrule
Cond. AR & 0:34 & -- & 0:35 & -- \\
Joint AR & 0:56 & -- & 1:23 & 0:15 \\
Joint PF & -- & 0:54 & -- & --\\
Joint PF (N=128) & -- & -- & -- & 2:48\\
Cond. PF & 0:43 & -- & 0:48 & --\\
EnJoi & 1:02 & -- & 1:32 & 0:17 \\
\bottomrule
\end{tabular}
\end{table}

% \section{Additional Visualizations}

\begin{figure}[!htb]
    \centering
    \includegraphics{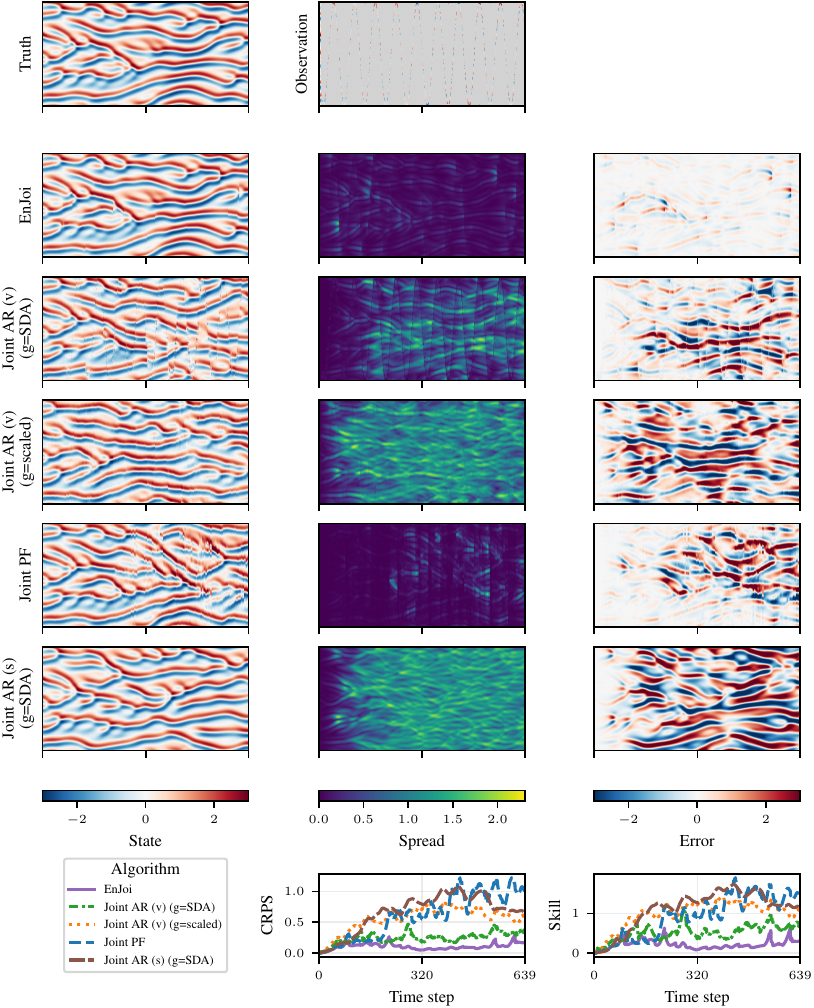}
    \caption{KS assimilation experiment with sine pattern observation. The left column displays one sample from the ensemble filters.}
    \label{fig:ks-errors}
\end{figure}

\begin{figure}[!htb]
    \centering
    \includegraphics{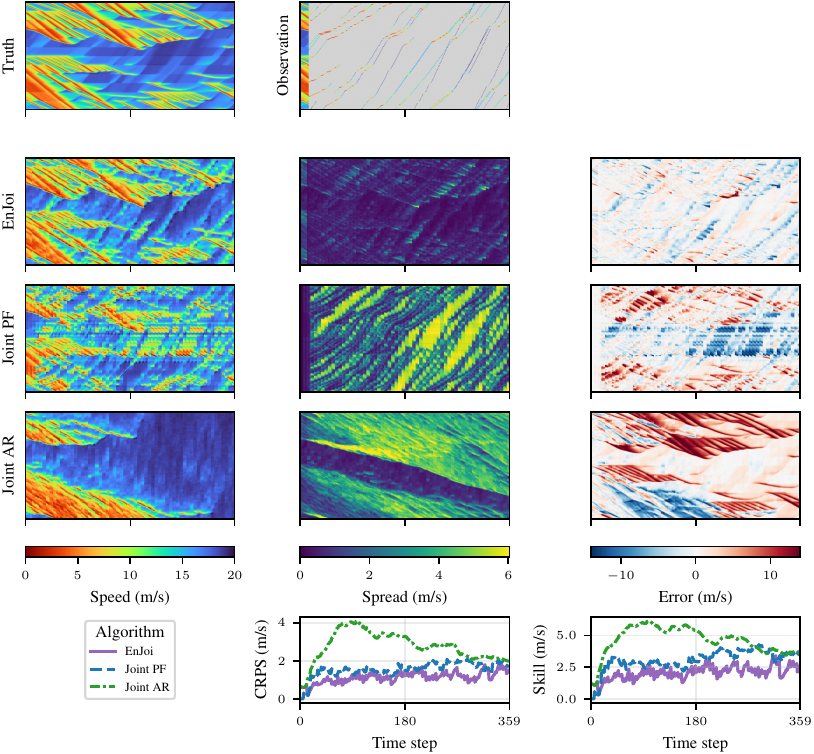}
    \caption{ARZ assimilation experiment. The left column displays one sample from the ensemble filters.}
    \label{fig:arz-errors}
\end{figure}

\end{document}

%% file: notations.tex
\definecolor{notationKo}{HTML}{1F77B4} % blue
\definecolor{notationKpo}{HTML}{D62728} % red
\definecolor{notationKkpo}{HTML}{9467BD} % purple
\definecolor{notationKt}{HTML}{1B7F24} % green
\definecolor{notationKpt}{HTML}{CC5500} % orange
\definecolor{notationKkpt}{HTML}{E6B800} % yellow
\definecolor{notationBase}{HTML}{4D4D4D} % black

\newcommand{\notationcolor}[2]{\textcolor{black}{#2}}

\newcommand{\nkt}[1]{\notationcolor{notationKt}{#1}}
\newcommand{\nkpt}[1]{\notationcolor{notationKpt}{#1}}

\newcommand{\nko}[1]{\notationcolor{notationKo}{#1}}
\newcommand{\nkpo}[1]{\notationcolor{notationKpo}{#1}}

\newcommand{\x}{\nko{\mathbf{x}}}
\newcommand{\xk}{\nko{\mathbf{x}^{k}}}
\newcommand{\xkp}{\nkpo{\mathbf{x}^{k+1}}}
\newcommand{\xkkp}{\nko{\mathbf{x}^{k}}\nkpo{^{:k+1}}}
\newcommand{\xt}{\nkt{\mathbf{x}_{t}}}
\newcommand{\xtm}{\nkt{\mathbf{x}_{t-1}}}
\newcommand{\xo}{\nko{\mathbf{x}_{0}}}
\newcommand{\xkt}{\nkt{\mathbf{x}^{k}_{t}}}
\newcommand{\xkpt}{\nkpt{\mathbf{x}^{k+1}_{t}}}
\newcommand{\xkkpt}{\nkt{\mathbf{x}}^{\nkt{k}\nkpt{:k+1}}_{\nkt{t}}}
\newcommand{\xkkptm}{\nkt{\mathbf{x}}^{\nkt{k}\nkpt{:k+1}}_{\nkt{t-\Delta t}}}
\newcommand{\xko}{\nko{\mathbf{x}^{k}_{0}}}
\newcommand{\xkpo}{\nkpo{\mathbf{x}^{k+1}_{0}}}

\newcommand{\xki}{\nko{\mathbf{x}^{k,(i)}}}
\newcommand{\xkpi}{\nkpo{\mathbf{x}^{k+1,(i)}}}

\newcommand{\y}{\nko{\mathbf{y}}}

\newcommand{\ykp}{\nkpo{\mathbf{y}^{k+1}}}
\newcommand{\ykkp}{\nko{\mathbf{y}^{k}}\nkpo{^{:k+1}}}

\newcommand{\yonek}{\nko{\mathbf{y}^{1:k}}}
\newcommand{\yonekp}{\nko{\mathbf{y}^{1}}\nkpo{^{:k+1}}}

\newcommand{\barx}{\nko{\mathbf{\bar{x}}}}
\newcommand{\barxk}{\nko{\mathbf{\bar{x}}^{k}}}
\newcommand{\barxkp}{\nkpo{\mathbf{\bar{x}}^{k+1}}}

\newcommand{\hatxk}{\nko{\mathbf{\hat{x}}^{k}}}
\newcommand{\hatxkp}{\nkpo{\mathbf{\hat{x}}^{k+1}}}

\newcommand{\checkxk}{\nko{\mathbf{\check{x}}^{k}}}

\newcommand{\checkxki}{\nko{\mathbf{\check{x}}^{k,(i)}}}

\newcommand{\tildexo}{\nko{\mathbf{\tilde{x}}_{0}}}

\newcommand{\tildexko}{\nko{\mathbf{\tilde{x}}^{k}_{0}}}
\newcommand{\tildexkpo}{\nkpo{\mathbf{\tilde{x}}^{k+1}_{0}}}
\newcommand{\tildexkkpo}{\nko{\mathbf{\tilde{x}}^{k}_{0}}\nkpo{^{:k+1}}}

\newcommand{\Pk}{\nko{\mathbf{P}^{k}}}
\newcommand{\Pkp}{\nkpo{\mathbf{P}^{k+1}}}